\documentclass[runningheads]{llncs}

\usepackage{eccv}

\usepackage{eccvabbrv}

\usepackage{graphicx}
\usepackage{booktabs}

\usepackage{graphics} % for pdf, bitmapped graphics files
\usepackage{epsfig} % for postscript graphics files
\usepackage{times} % assumes new font selection scheme installed
\usepackage{amsmath} % assumes amsmath package installed
\usepackage{amssymb}  % assumes amsmath package installed

\usepackage{comment}

\usepackage{booktabs}
\usepackage{pifont}% http://ctan.org/pkg/pifont
\usepackage{algorithm}
\usepackage{algpseudocode}
\usepackage{graphicx}
\usepackage{float}
\usepackage{tabularx,colortbl}
\usepackage{xparse,mathtools}
\usepackage{diagbox}
\usepackage{adjustbox,lipsum}
\usepackage{breqn}
\usepackage{wrapfig}
\mathchardef\mhyphen="2D
\usepackage[utf8]{inputenc}

\newcommand{\cmark}{\ding{51}}%
\newcommand{\xmark}{\ding{55}}%

\usepackage{tabulary,overpic}

\newlength\savewidth

\usepackage{bbm}
\usepackage{xspace}
\usepackage{dsfont}
\usepackage{multirow}
\newcommand{\ba}{\mathbf{a}}

\newcommand{\be}{\mathbf{e}}
\newcommand{\bI}{\mathbf{I}}

\newcommand{\bo}{\mathbf{o}}
\newcommand{\bp}{\mathbf{p}}

\newcommand{\bs}{\mathbf{s}}

\newcommand{\btheta}{\boldsymbol{\theta}}\newcommand{\bTheta}{\boldsymbol{\Theta}}

\newcommand{\bphi}{\boldsymbol{\phi}}

\newcommand{\bomega}{\boldsymbol{\omega}}

\newcommand{\cA}{\mathcal{A}}

\newcommand{\cD}{\mathcal{D}}

\newcommand{\cH}{\mathcal{H}}

\newcommand{\cL}{\mathcal{L}}

\newcommand{\cO}{\mathcal{O}}

\newcommand{\cS}{\mathcal{S}}

\makeatletter
\DeclareRobustCommand\onedot{\futurelet\@let@token\@onedot}
\def\@onedot{\ifx\@let@token.\else.\null\fi\xspace}
\def\eg{e.g\onedot} 
\def\ie{i.e\onedot} 
\def\cf{cf\onedot} 
\def\etc{etc\onedot}

\def\etal{et~al\onedot}

\makeatother

\newcommand{\boldparagraph}[1]{\vspace{0.2cm}\noindent{\bf #1:}}

\definecolor{darkgreen}{rgb}{0,0.7,0}

\definecolor{blue}{HTML}{49a0d4}
\definecolor{red}{HTML}{cc1100}
\definecolor{orange}{HTML}{cc7700}
\definecolor{gray}{HTML}{efefef}
\definecolor{darkgreen}{HTML}{228B22}
\definecolor{darkgray}{HTML}{757575}

\usepackage[accsupp]{axessibility}  % Improves PDF readability for those with disabilities.

\usepackage[breaklinks,colorlinks,citecolor=eccvblue]{hyperref}
\usepackage{orcidlink}

\usepackage[capitalize]{cleveref}
\crefname{section}{Sec.}{Secs.}
\Crefname{section}{Section}{Sections}
\Crefname{table}{Table}{Tables}
\crefname{table}{Tab.}{Tabs.}

\begin{document}

% ---------------------------------------------------------------
% TODO REVIEW: Replace with your title

%\def\MethodName{\textbf{\textit{Lasagna}}}
%Cloud Residual Policies
%Situational Pedestrian Navigation via Cloud-Local Collaboration
%CRISP
%OffLoader
% \title{Situational Offloading for Cloud-based Vision-based Policies}
%Highlight novlety. Residual. 
%Unified Cloud-Local
%CRISP
%Local-Cloud Decision-Making for Vision-Based Policies
%ECO
 % \title{Situational Cloud-Local Policies} 
% \title{UniLC: Unifying Local and Cloud Computation for Decision-Making Policies}
%Highlight efficency. 
%UniLCD: 
%Cloud-based Policy Adaptation via RL
%Eshed: Don't do this, create seperate file. 
\title{
%Supplementary Material for \\ 
Unified Local-Cloud Decision-Making via \\ Reinforcement Learning
}

% TODO REVIEW: If the paper title is too long for the running head, you can set
% an abbreviated paper title here. If not, comment out.
\titlerunning{Unified Local-Cloud Decision-Making via Reinforcement Learning}

% TODO FINAL: Replace with your author list. 
% Include the authors' OCRID for the camera-ready version, if at all possible.
% \author{First Author\inst{1}\orcidlink{0000-1111-2222-3333} \and
% Second Author\inst{2,3}\orcidlink{1111-2222-3333-4444} \and
% Third Author\inst{3}\orcidlink{2222--3333-4444-5555}}

% % TODO FINAL: Replace with an abbreviated list of authors.
% \authorrunning{F.~Author et al.}
% % First names are abbreviated in the running head.
% % If there are more than two authors, 'et al.' is used.

\author{Kathakoli Sengupta \and
Zhongkai Shangguan \and Sandesh Bharadwaj \and Sanjay Arora$^{\dagger}$ \and \\ Eshed Ohn-Bar \and Renato Mancuso
}

% % TODO FINAL: Replace with your institution list.
% \institute{Princeton University, Princeton NJ 08544, USA \and
% Springer Heidelberg, Tiergartenstr.~17, 69121 Heidelberg, Germany
% \email{lncs@springer.com}\\
% \url{http://www.springer.com/gp/computer-science/lncs} \and
% ABC Institute, Rupert-Karls-University Heidelberg, Heidelberg, Germany\\
% \email{\{abc,lncs\}@uni-heidelberg.de}}

% \author{Kathakoli S \and
% Jimuyang Zhang\inst{*} \and
% Eshed Ohn-Bar}
% TODO FINAL: Replace with an abbreviated list of authors.
% \autor{Kathakoli Sengupta \and Zhongkai Shagguan \and Sandesh Bharadwaj \and Sanjay Arora \and Eshed Ohn-Bar \and Renato Mancuso}}

\authorrunning{Sengupta et al.}
% First names are abbreviated in the running head.
% If there are more than two authors, 'et al.' is used.

% TODO FINAL: Replace with your institution list.
% \institute{Princeton University, Princeton NJ 08544, USA \and
% Springer Heidelberg, Tiergartenstr.~17, 69121 Heidelberg, Germany
% \email{lncs@springer.com}\\
% \institute{Boston University, Boston MA 02215, USA \\
% \email{\{huangtom,zhangjim,eohnbar\}@bu.edu}}

\institute{Boston University \hspace{0.3cm} $^{\dagger}$Red Hat}
% \email{\{huangtom,zhangjim,eohnbar\}@bu.edu}}

\maketitle
\begin{abstract}
    Embodied vision-based real-world systems, such as mobile robots, require a careful balance between energy consumption, compute latency, and safety constraints to optimize operation across dynamic tasks and contexts. As local computation tends to be restricted, offloading the computation, \ie, to a remote server, can save local resources while providing access to high-quality predictions from powerful and large models. However, the resulting communication and latency overhead has led to limited usability of cloud models in dynamic, safety-critical, real-time settings. To effectively address this trade-off, we introduce UniLCD, a novel hybrid inference framework for enabling flexible local-cloud collaboration. By efficiently optimizing a flexible routing module via reinforcement learning and a suitable multi-task objective, UniLCD is specifically designed to support the multiple constraints of safety-critical end-to-end mobile systems. We validate the proposed approach using a challenging, crowded navigation task requiring frequent and timely switching between local and cloud operations. UniLCD demonstrates improved overall performance and efficiency, by over 35\% compared to state-of-the-art baselines based on various split computing and early exit strategies. Our code is available at \url{https://unilcd.github.io/}.
\end{abstract}
%test\cite
\section{Introduction}
\label{sec:intro}

We are currently undergoing a transformative societal phase as vision-based systems, such as mobile robots and personalized devices, transition from their controlled lab environments and into the real world. However, computational and energy constraints currently hinder the performance of deployed real-world systems, \eg, high-cost systems today may only be able to support lightweight neural network models with limited accuracy and runtime of about an hour before the on-board battery is depleted~\cite{hadidi2019characterizing,kang2017neurosurgeon,fu2021minimizing,miki2022learning,liu2024enhanced,zhuang2023robot,song_going_2023}. Current efficiency bottlenecks may only become worse over time given ever-increasing sensor resolutions
%, higher-dimensional inputs, 
and larger models~\cite{brohan2023rt,shah2023lm,lin2023video,liu2024visual,achiam2023gpt,touvron2023llama}. 

To address the need for reliable and accurate inference, the currently prevailing approach for processing with increasingly powerful models involves sending data, \eg, an image, to remote cloud servers in order to offload computation~\cite{li2020energy,zhang2021energy,achiam2023gpt,brohan2023rt,kang2017neurosurgeon,lin2019cost}. While this practice can benefit real-world systems through delivering high-quality model predictions, the transmission and cloud processing overhead incurs a high latency cost and is therefore not suitable for dynamic real-world agents, such as mobile systems, which must continuously anticipate and respond to dynamic settings. Safety-critical systems with hard real-time constraints and responsiveness may solely rely on local processing, while leveraging strategies such as model pruning~\cite{jiang2022model, gamanayake2020cluster, wu2023model, kong2021spvit} and quantization~\cite{wang2021lightweight, kryzhanovskiy2021qpp, polino2018model} to support embodied constraints, \eg, limited hardware and battery life. However, on-device lightweight models often suffer from significantly degraded accuracy and in turn can hinder safe decision-making, \ie, missing or falsely detecting nearby pedestrian. Thus, we seek to address a fundamental research question: How to realize robust vision-based systems that can be flexibly optimized for both \textit{safety and real-time efficiency} while operating in dynamic real-world settings? 

While shared local-cloud computation frameworks have been extensively studied by prior works, \eg, for various Internet of Things applications~\cite{gan2023cloud,kag2022efficient,zhang2021energy,li2020energy,hu2020coedge,park2021collaborative,hu2019edge,kang2017neurosurgeon,kag2022efficient,song_looking_2024}, the aforementioned methods rarely analyze safety-critical, real-time operation. Instead, standard off-loading frameworks rely on various ad-hoc and inadequate heuristics, \eg, based on a task-agnostic layer-wise split or simplistic statistics~\cite{kang2017neurosurgeon,kag2022efficient}, and thus cannot accommodate dynamic inference, \ie, based on the difficulty or safety-constraints of the current scenario. For instance, some scenarios may require split-second decisions, such as in the case of dense and dynamic scenarios with crowded surrounding pedestrians. In this work, we develop a more generalized paradigm enabling situation-specific collaboration between cloud computing and local inference to balance the energy cost while maintaining high system accuracy. Our method can be used to flexibly optimize for multiple constraints, \ie, latency, accuracy, efficiency, and safety, while also enabling systems with a highly sustainable operation (a critical need as compute-intensive systems become more widely adopted~\cite{sudhakar2022data}). 

\boldparagraph{Contributions} 
Towards facilitating real-time, cloud-enabled systems, we make three key contributions. First, we introduce a novel cloud-local hybrid collaboration framework, UniLCD. Our proposed reinforcement learning-based framework is inspired by the observation that small, low-power consumption models on edge devices are sufficient in some scenarios, while more complex conditions may call for powerful cloud resources. Our approach employs a conditional routing module that learns to dynamically route computation between local and cloud resources. Second, we demonstrate the importance of carefully designing the reward function in order to effectively balance over multiple constraints, \ie, energy consumption, latency, and safe performance in a context-dependent manner. Third, we validate our method and several state-of-the-art baselines by introducing a novel benchmark comprising challenging vision-based crowded navigation tasks that require seamless decision-making with frequent cloud-local switching. We demonstrate our approach to significantly outperform prior methods~\cite{kang2017neurosurgeon,kag2022efficient} by 35\% across multiple metrics, as well as a proposed ecological score. Our code and benchmark are available at \url{https://unilcd.github.io/} for future researchers tackling multifaceted challenges in practical, real-world, vision-based systems. 

\section{Related Work}
\label{sec:related}

% \Kathakoli{Modify this table}
\boldparagraph{Cloud-Edge Collaborative Systems}
The integration of cloud computing and edge devices has attracted increasing attention in recent years as it can potentially combine the advantages from both~\cite{ding2020dynamic, satyanarayanan2021role, penmetcha2021deep, gan2023cloud}. 
Leveraging the cloud's capacity to employ advanced hardware and deploy large models allows quick execution of computationally intensive tasks with accurate results, while edge devices enable real-time processing and diminish latency by accessing data close to the source.
Thus, cloud-edge collaboration systems have been developed across various domains, such as autonomous driving~\cite{kim2021edge} and smart cities~\cite{wu2020collaborate}. Recently, Kag~\etal~\cite{kag2022efficient} proposed a hybrid approach that learns to efficiently select queries for cloud processing. However, the method focuses on simplistic image classification domains without integrating sensitivity to computational delays. In contrast, our framework emphasizes the importance of \textit{both prompt and accurate} cloud-edge collaboration and can thus more flexibly support broad applications, \eg, with handling of dynamic and complex safety-critical scenarios.  

\begin{table}[!t]
\caption{\textbf{Comparison with Prior Work.}
% \red{This is not really the way to highlight our method. bolded title is too long, just make to the point. "Comparison with Related Prior Work.}
By holistically tackling aspects of system safety, efficiency, and latency, our framework is suitable for real-time decision-making in dynamic scenes. 
% and complex complex and 
%We emphasize that our settings are more complex compared to prior works which do not generally include 
% \red{We are contextual/situational, I think selective query is not.\Kathakoli{Need to add safety to this table}}
} 

  \centering
  \resizebox{\columnwidth}{!}{%
  \begin{tabular}{l|ccccccc}
    \hline
    \multirow{2}{*}{\textbf{Method}} & \textbf{Low} & \textbf{Edge} & \textbf{End-to-End} & \textbf{\hspace{0.1cm} Situational \hspace{0.1cm}} & \textbf{Embedding} & \textbf{High} & \textbf{Real}\\
    & \textbf{Latency} & \textbf{Deployment} & \textbf{Training} & & & \textbf{Accuracy} & \textbf{Time}\\
    \hline
    On-Device & \cmark & \cmark & - & \cmark & \xmark & \xmark  & \cmark \\
    On-Cloud & \xmark & \xmark & - & \cmark & \xmark & \cmark  & \xmark \\
    Neurosurgeon~\cite{kang2017neurosurgeon} & \xmark & \cmark & \cmark & \cmark & \cmark & \cmark  & \xmark\\
    Dynamic~\cite{han2021dynamic} & \cmark & \cmark & \xmark & \xmark & \cmark & \xmark  & \xmark\\
    Selective Query~\cite{kag2022efficient} & \cmark & \cmark & \cmark & \xmark & \xmark & \cmark  & \xmark\\
    Compressive Offloading~\cite{yao2020deep} & \xmark & \xmark & \cmark & \cmark & \cmark & \cmark  & \xmark\\
    Adaptive Offloading~\cite{van2021adaptive} & \cmark & \cmark & \cmark & \cmark & \xmark & \xmark  & \xmark\\
    Deep Sequential RL~\cite{wang2019computation} & \cmark & \cmark & \cmark & \cmark & \cmark & \xmark  & \xmark\\
    \rowcolor{gray}
    UniLCD (ours) & \cmark & \cmark & \cmark & \cmark & \cmark & \cmark  & \cmark\\
    \hline
  \end{tabular}
  }
  \label{tab:checkmarks}
  % \vspace{-1em}
\end{table}

\boldparagraph{Real-Time Decision-Making}
Real-time decision-making is critical for scenarios that require fast and accurate responses (\eg, autonomous driving~\cite{lai2023xvo,lin2019deep,zhang2024feedback,lai2024uncertainty}, energy grid management~\cite{wu2018deep,shangguan2021neural}, assistive technologies~\cite{treuillet2010outdoor,huang2022assister,dang2024real}). State-of-the-art models usually have billions of parameters, resulting in long inference times. Notably, on-going work is primarily concentrated on enhancing data and model processing speed~\cite{frantar2022spdy, wang2019haq}, as well as designing dedicated processing chips~\cite{karras2020hardware, shao2019simba}.
In this work, we tackle a complementary direction of cloud-edge collaboration, where we explore the integration of the cloud's accelerated processing speeds with the deployment of edge devices. This integration aims to effectively mitigate latency concerns while ensuring the reliability of prompt decision-making, which generally results in efficiency and latency costs~\cite{park2020adaptive, masoudi2020device, aujla2017optimal}.

\boldparagraph{Early-Exit Models}
Early exit techniques are closely related to our work as they incorporate internal classifiers at various shallow layers, allowing the model to exit earlier during inference while still predicting the correct label, \ie, to save time and energy costs~\cite{laskaridis2021adaptive,qendro2021early,tang2023you,li2021appealnet,xia2023window,jazbec2024fast,kang2017neurosurgeon}. Yet, despite the fast inference, models can still suffer in latency and accuracy~\cite{kang2017neurosurgeon}.
Li~\etal introduced AppealNet~\cite{li2021appealnet}, an architecture that efficiently processes deep learning tasks by predicting whether inputs can be managed by a resource-constrained edge device or need to be offloaded to a cloud-based model. 
However, such early exit strategies only addresses on-device efficiency without considering overall task and safety, 
% especially for time-sensitive tasks, 
leading to potentially unsafe and frequent navigation errors in our robot navigation task, as will be shown in our analysis in Sec~\ref{sec:analysis}.

\boldparagraph{Energy-Efficient Models}
Our proposed method primarily aims to present timely results while minimizing the energy cost from both cloud servers and edge devices, which can not only help reduce carbon emissions~\cite{kimovski2021cloud, alharbi2021energy, zhu2022psto, chin2021pareco} but also increase battery life and reduce total overhead costs~\cite{lin2019cost, el2018computational,hadidi2019characterizing}.
The promise of energy-efficient models has been substantially developed through various research fields, focusing on optimizing computational efficiency and resource utilization.
One aspect of the field of energy-efficient models emphasizes hardware design~\cite{judd2016stripes, nguyen2019high}. However, hardware development is often time-consuming and expensive.
Another aspect provides insight into the efficacy of model pruning and quantization techniques that can minimize model size thus helping save energy~\cite{yang2017designing,zhu2017target,jin2024learning} consumption.
In our work, we delve into resource allocation and optimization strategies, seeking to further improve energy efficiency.

\boldparagraph{Decision-Making Models for Mobile Systems}
Robotic companions and assistants designed to operate in human-inhabited environments has made significant progress~\cite{katyal2020intent, chen2019crowd,xworld}.
Imitation learning~\cite{hussein2017imitation,zhang2022selfd,zhang2023coaching,zhu2023learning} involves the system learning from observed behavior, typically demonstrated by a human or professional operator, to guide its decision-making process in mobile systems, such as autonomous vehicles~\cite{pan2017agile,teng2022hierarchical} or drones~\cite{wang2021robust}.
Reinforcement learning (RL)~\cite{wiering2012reinforcement,yang2019deep,lin2020hpt,moerland2023model,zheng2023coordinated} has also been applied to optimize decision-making processes, allowing robots to navigate efficiently while avoiding obstacles and optimizing path planning~\cite{wang2019computation,caruso2023robot,zhu2021deep}.
However, the wide-scale deployment of real-time vision-based navigation models on robots today is hindered by ever-increasing computational and hardware constraints~\cite{li2023red,dudek2024computational}, making it difficult to deploy accurate but large models on systems. 
In our work, our RL-based framework can be used to query the cloud efficiently, \ie, to reduce overall energy cost without sacrificing local resources ~\cite{hanyao2021edge,liu2019edge,yao2020deep,wang2020fast}. Our flexible framework is model and platform-agnostic, while automatically adapting across situations and communication settings. 

\section{Method}
\label{sec:method}
\begin{figure}[!t]
    \centering
    \includegraphics[trim={5cm 2cm 8cm 4cm},clip,width=0.95\textwidth]{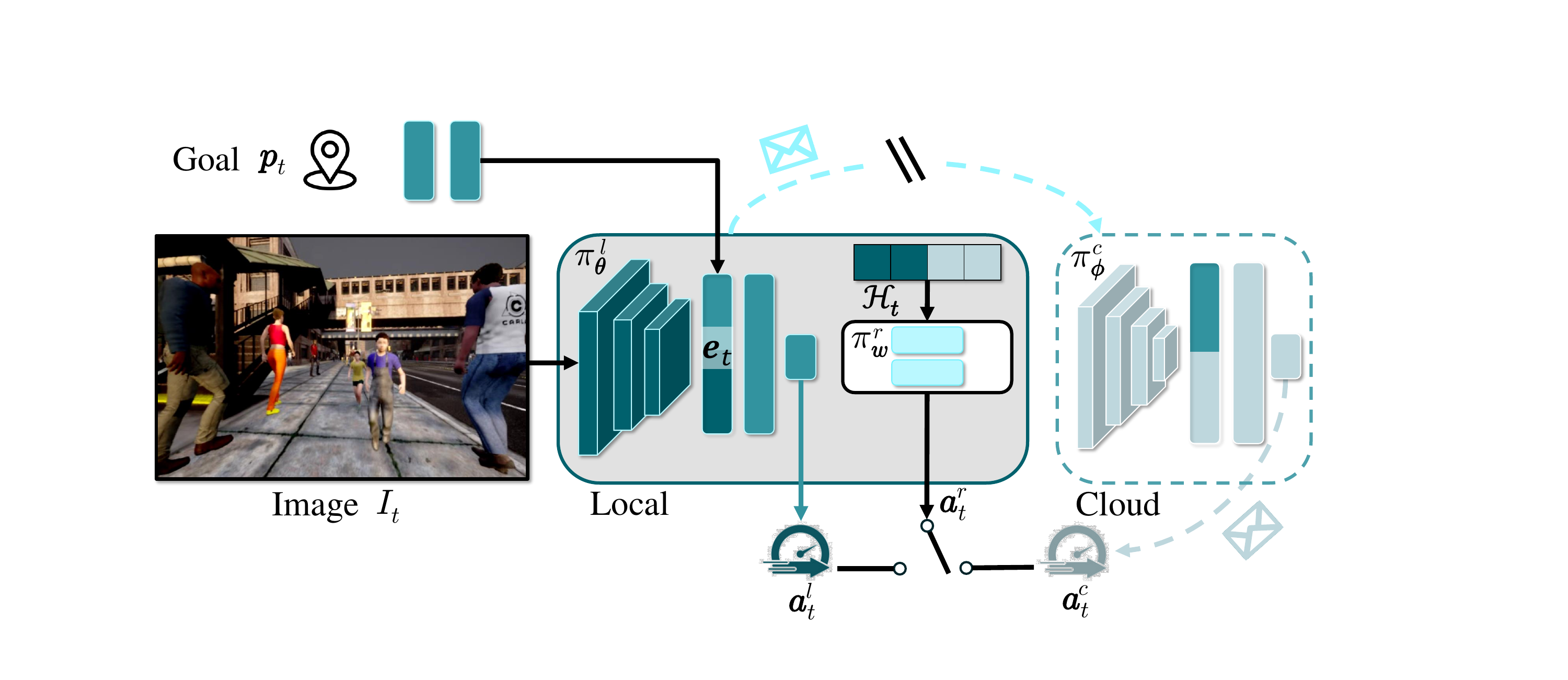} 
    \vspace{-0.3cm}
    % Adjust width as necessary for two columns
    \caption{ \textbf{Overview of UniLCD.} Our system comprises a situational routing module, which takes the current embedding and a history of previous actions. Local actions are predicted by a pre-trained lightweight model that can be deployed efficiently on a the mobile system. The sample-efficient routing module, trained via RL, determines whether to implement the local action or transmit the scene embedding to the cloud server model, which is more accurate but computationally expensive and induces latency.
    }
    \label{fig:architecture}
\end{figure}

Our objective is to learn a \textit{policy} that determines the dynamic allocation of local and cloud resources, optimizing energy efficiency and real-time performance. In the following, we discuss the main components of our method. First, we formulate our real-time task of navigating to a specific goal in Sec.~\ref{subsec:form}.
Next, we employ imitation learning to train two navigation policies tailored for local and cloud settings, respectively (Sec.~\ref{subsec:model}). Third, we discuss how we train our sample-efficient \textit{routing policy} via Proximal Policy Optimization (PPO)~\cite{schulman2017proximal} (Sec.~\ref{subsec:policy}). In particular, we introduce our multi-objective reward, designed to optimize energy efficiency while maintaining navigation performance. 
\cref{fig:architecture} depicts an overview of our method.

\subsection{Problem Formulation}
\label{subsec:form}
We formulate our robot navigation task as a real-time sequential decision-making problem from a set of observations $\bo=\{\bI, \bp\} \in \cO$, comprising a front-view camera image $\bI \in \mathbb{R}^{W\times H \times 3}$, the next waypoint $\bp \in \mathbb{R}^{2}$,
to a set of actions $\ba=\{d, v\} \in \cA$, where $d \in \mathbb{R}$ represents the orientation and $v \in \mathbb{R}$ is the speed~\cite{codevilla2018end,dosovitskiy2017carla}.
Our objective is to obtain a mapping function $f_{\bTheta} \colon \cO \to \cA$ parameterized by $\bTheta = [\btheta, \bphi, \bomega]
$ for generating actions at each time step to minimize the overall energy cost from both the local device and cloud server~\cite{chen2021energy}. 
Our approach consists of three policies: a local and a cloud navigation policy $\pi_\btheta^{l}$ and $\pi_\bphi^{c}$ with weights $[\btheta, \bphi]$ that map the observations to actions, producing $\ba_t^l$ and $\ba_t^c$,
respectively. Additionally,  we introduce a routing policy $\pi_\bomega^r(\bo | \pi_\btheta^{l}, \pi_\bphi^{c})$
parameterized by $\bomega$, to dynamically determine the optimal utilization of local resources versus offloading to the cloud server.

While our method can be optimized end-to-end, we initialize the two navigation policies in an imitation learning manner~\cite{codevilla2018end, hawke2020urban, pan2020imitation}. We also note that, given the constraints of local device hardware in our setting, the number of parameters for the local navigation policy is generally smaller than that of the cloud navigation policy.

For comparative analysis and sample efficiency, after training the navigation policies, we freeze the parameters $\btheta, \bphi$ and train the routing policy $\pi_\bomega^r(\bo | \pi_\btheta^{l}, \pi_\bphi^{c})$ based on the observations and the two navigation policies using residual RL.
It's noteworthy that both the local navigation policy and the routing policy are deployed on a local device, sharing parameters for the initial layers of the neural network, as shown in~\cref{fig:architecture}.

\boldparagraph{Cloud-Aware Robot Navigation Task}
Due to the challenges of obtaining sufficient data in the dynamic real-world scenario~\cite{mavrogiannis2023core, tsai2020generative, chung2010change}, we develop a simulation environment tailored for robot navigation through a crowded outdoor setting. 
Our environment is built upon CARLA (version 0.9.13)~\cite{dosovitskiy2017carla}, an open-source simulator typically employed for testing autonomous driving algorithms. 
The information transmission between the local and the cloud server potentially introduces a stochastic delay~\cite{blog, hooghiemstra2001delay, kang2017neurosurgeon, zhang2007modeling}. 

\boldparagraph{Toward Energy-Efficiency}
The primary objective of our research is to enhance energy efficiency while preserving task performance~\cite{kang2017neurosurgeon,hu2020coedge,park2021collaborative}.  
In our study, the navigation task is offloaded to the cloud for processing only when local computation is insufficient for achieving optimal task performance. However, communication bottlenecks necessitate prioritizing local processing to conserve energy and minimize latency. Offloading to the cloud is reserved for situations where only the cloud can meet the task performance requirements, such as challenging scenarios like navigating in bad weather~\cite{mușat2021multi} or crowded environments.
Consequently, we leverage a routing network trained with PPO and suitable for running on local devices. This modular architecture is then optimized for energy efficiency, as discussed below. 

\begin{algorithm}[!t]
\begin{minipage}{\textwidth}
% \small
  \caption{UniLCD's Routing Policy Training with Reinforcement Learning } 
  \label{alg:switch}
  \begin{algorithmic}[1]
    \State \textbf{Input:} Image $\bI$, next waypoint $\bp$, local policy $\pi_\btheta^l$, cloud policy $\pi_\bphi^c$ 
    % \Comment{$\bI$: image, $\bp$: current position} 
    \State \textbf{Initialize:} Number of iterations $T$, history $\cH$, routing policy $\pi^r_{\bomega}$, reply buffer $\cS$
    \State Collect on policy samples:
    \For{$\text{t}=1$ \textbf{to} $T$}
    \State Obtain local action $\ba_t^l$ and embeddings $\be_t$ using local policy $\pi_\btheta^l (\bI_{t}, \bp_t)$
    \State Append $(\ba_t^l, 0)$ to history $\cH_t$ %and remove the first value
    \If{$\pi^r_{\bomega_t}(\cH_t, \be_t)=0$} $\ba_t=\ba_t^l$
    \Else 
    \State Send $\be_t$ to cloud, $\ba_t = \pi_\bphi^c (\bI_t, \bp_t)$  
    \State Update last value of $\cH_t$ to $(\ba_t, 1)$
    \EndIf
    \State Compute instant reward using \cref{eq:reward} %$r_t$
    \If{Arrived destination} break \EndIf
    \State Update replay buffer $\cS=\cS \cup \{\bI_t, \bp_t, \cH_t, r_t\}$
    %[\ref{eq:reward}]
    \State Update routing policy parameters with PPO
    \EndFor
  \end{algorithmic}
  % \vspace{0.1cm}
\end{minipage}
\end{algorithm}

\begin{comment}
\end{comment}

% Learning robot Navigation Policy
\subsection{Learning Local and Cloud Policies}
\label{subsec:model}
In our work, we follow the standard imitation learning approach to train our navigation policies in an offline manner.
Specifically, we first collect a dataset $\cD=\{\bI_i, \bp_i, \ba_i\}_{i=1}^N$ on diverse and complex routes and weathers using CARLA~\cite{dosovitskiy2017carla,ohn2020learning} to simulate real-world scenarios to provide the basis for our navigation policies.

As shown in~\cref{fig:architecture}, both the local and cloud navigation policies comprise (i) a visual semantics feature extractor module for obtaining embeddings from input images, (ii) a multilayer perceptron to extract the feature associated with the robot's imminent goal point, and (iii) a goal-conditional module that takes concatenation of image embeddings and goal features to predict robot actions, encompassing both direction $d$ and speed $v$. 
To share common features and accommodate local computing resource constraints, we first train a robust cloud policy with a large neural network. 
Subsequently, we freeze the parameters of the first few layers of the pre-trained cloud policy to serve as a shared feature extractor for the local policy. Additional fully connected layers are then added to the extracted features to form the local policy. 
Therefore, the local policy is significantly smaller compared to the cloud policy.
Only the parameters of the fully connected layer in the local policy are optimized, leading to reduced computational consumption and improved efficiency in both training and inference time.

The learning objective for both local and cloud policies is achieved by minimizing the $\cL_1$ distance: 
\begin{equation}
    \underset{}{\text{minimize}} \, \mathbb{E}_{(\bI,  \bp, \ba) \sim D} [\cL_1(\ba, \pi(\bI, \bp))]
\end{equation}
where $\pi$ represents navigation policy $\{\pi_\btheta^l, \pi_\bphi^c\}$.

\subsection{Learning a Routing Policy}
\label{subsec:policy}
To effectively balance cloud and local computing resources while achieving good performance at the same time, our study proposes a \textit{routing policy} that seamlessly switches between local and cloud. We follow the standard PPO training process as shown in~\cref{alg:switch} and the formulation of our routing policy is introduced below.

\boldparagraph{State Space} 
We denoted the current state as $\bs_t=\{\be_t, \cH_t\}$, where $\be_t$ represents the image embeddings and goal features extracted from the shared feature extractor employed by both local and cloud policies, as discussed in Sec.~\ref{subsec:model}, and $\cH_t = {\{(\ba_i, \mathds{1}_i})\}_{t-k}^{t} $ is a sequence of the last $k$ history actions, where $\ba_i$ represents previous navigation actions and $\mathds{1}_i$ is an indicator function indicating whether the action is obtained locally or from the cloud. 

\boldparagraph{Action Space} 
The action produced by our router policy is a binary discrete value, indicating whether to accept the local navigation policy $\pi_\btheta^l$ or transmit the embeddings $\be_t$ to the cloud navigation policy $\pi_\bphi^c$.

\boldparagraph{Task Reward}
It is challenging to optimize task performance and energy efficiency while considering the sub-optimal actions generated from a small local model and latency-induced cloud model.

We design our reward function encompassing five key components: geodesic reward $r_{geo}$, speed reward $r_{speed}$, energy disadvantage bonus $r_{energy}$, extreme action clip $r_{action}$, and collision penalty $r_{collision}$. The overall reward function is defined as 
\begin{equation}\label{eq:reward}
r = \left( r_{geo} \cdot r_{speed} \cdot r_{energy} \cdot r_{action} \right)^{\alpha}-r_{collision}
\end{equation} 
where $\alpha=1/4$ is a scaling factor that scales the overall reward between $[0, 1]$. By using a multiplicative objective, we ease optimization across different instantaneous components of the reward, \ie, if one of the terms is low, the entire reward is affected. Hence, the multiplicative overall reward objective can also be scale invariant to some extent (however, clipping one of the reward terms can change its importance). In our analysis, we find it to improve optimization and reduce the need for careful hyper-parameter tuning in multi-objective RL. For collision reward, we treat the term separately as it is sparsely observed over selected frames as a large negative reward, as defined below. 

\boldparagraph{Geodesic Reward}
We introduce $r_{geo}$ to align the robot's trajectory with the pre-defined path, penalizing the robot for movements that deviate from the path.
\begin{equation}
    \begin{aligned}
        r_{geo}=(1 - \tanh(d_{geo}))
    \end{aligned}
\end{equation}
where $d_{geo}$ represents the Euclidean distance from the robot's current position to the nearest waypoint on the pre-defined path.

\begin{comment}
\end{comment}

\boldparagraph{Speed Reward}
The speed reward, denoted as $r_{speed}$, is designed to motivate robots to reach their destination in the shortest time.
Considering the impact of GPU resource utilization on task execution duration, speed reward encourages quick task completion thus minimizing the overall computing resource usage. 
The maximum speed of a robot in our environment is set to $m_v = 1.5m/s$ according to~\cite{mohler2007visual, levine1999pace}, and we define the speed reward as 
\begin{equation}
\begin{aligned}
r_{speed}= v/m_v
\end{aligned}
\end{equation}
where $v$ represents the current speed.

\begin{comment}

\end{comment}

\boldparagraph{Energy Disadvantage}
In our pursuit of minimizing energy consumption as the primary objective, and considering the significantly lower energy cost of local computation compared to cloud transmission and computation, we introduce penalties when the robot seeks cloud assistance. 
Similar to the speed component, we formulate the normalized energy disadvantage reward as 
\begin{equation}
    \begin{aligned}
        r_{energy}= 1-e/m_e
    \end{aligned}
\end{equation}
where $m_e$ is the maximum energy cost at one step.

\boldparagraph{Extreme Action Clip}
Due to the property that $tanh$ converges at extreme action values, leading to a potential risk of converging to local maxima, we introduce an action clipping mechanism to address this concern. Specifically, actions exceeding the maximum allowed value will be directly assigned a reward of $0$. This precautionary measure is extended to both standardized directional and speed actions, as shown in~\cref{equ:ra}
\begin{equation}
\label{equ:ra}
    \begin{aligned}
        r_{action} = \mathds{1}(|r_{speed}|<\epsilon)\cdot  \mathds{1}(|\frac{d}{d_m}|<\epsilon)
    \end{aligned}
\end{equation}
where $\epsilon=0.97$ is the threshold for clipping the actions, $d$ is the next rotation angle suggested by the navigation model chosen and  $d_m$ is the maximum possible rotation, ensuring that extreme actions do not unduly influence the reward function.

\boldparagraph{Collision Disadvantage}
Robot navigation needs to emphasize safety and endeavor to avoid collisions with obstacles, ensuring the successful accomplishment to the destination.
To underscore the significance of collision avoidance, we include an episodic termination mechanism to emphasize the severity of collisions. Specifically, a substantially higher negative penalty $r_{collision}$ is assigned and the episode is terminated whenever a collision happens.

\section{Experiments}
\label{sec:analysis}

\subsection{Implementation Details}
\label{subsec:implementation}
\boldparagraph{Data Collection}
To learn the local and cloud navigation policies using imitation learning, we collect crowd navigation data from our CARLA environment. We simulate environments with varying densities by introducing five, 15, 30, and 70 pedestrians into the ego robot's path, corresponding to low, medium, dense, and crowd-density settings, respectively. This data is then aggregated to form a comprehensive training dataset for both local and cloud navigation policies. Our data collection approach employs a heuristic policy: we establish 10 different paths with predefined waypoints, which the robot follows from the starting position to the destination. The robot avoids static obstacles and halts when other pedestrians block its path, resuming moving once the path is clear.

\boldparagraph{Local and Cloud Policies}
As discussed in~\cref{subsec:model}, our local and cloud policies share a common feature extractor that processes a $480\times480$ image and a 2D imminent position vector. Specifically, we utilize the initial layers of RegNet~\cite{schneider2017regnet} to extract the image features, and two fully connected (FC) layers to capture the position features. For the local policy, the image feature is flattened and concatenated with the position feature, followed by additional FC layers to generate the local actions. 
For the cloud policy, both the image feature and position feature are transmitted from the local to the cloud. The image feature is then processed by the remaining layers of RegNet~\cite{schneider2017regnet} and concatenated with the position feature. An MLP is built on top of the combined features to generate the cloud actions.

\boldparagraph{Routing Policy}
Our routing policy follows the standard PPO training process, with a policy network that decides whether to route locally or to the cloud, and a value network that evaluates the chosen actions by estimating the value function. Both the policy and value networks are MLPs with two hidden layers, with the hidden size being 16 for the policy network and 256 for the value network.

\boldparagraph{Training Protocol}
We use AdamW~\cite{loshchilov2017decoupled} optimizer and train for 200 epochs with a learning rate of $0.0001$ using our robot navigation dataset for both local and cloud navigation policies.
The routing policy is trained for $1{,}000$ episodes, each consisting at most $1{,}500$ steps, with a discount factor $\gamma$ set to 0.99. Episodes are truncated if the agent collides with other objects, completes the designated number of steps, reaches the predefined destination, or deviates more than three meters from the predefined route.

\subsection{Evaluation Metrics}

\boldparagraph{Task Performance Evaluation}
We adapt CARLA's evaluation metrics~\cite{carlaleaderboard} to assess the navigation performance of our ego robot. In addition to commonly used metrics (\eg, success rate, route completion, etc.), we propose \textit{Navigation Score}, denoted as $NS$, similar to the driving score~\cite{carlaleaderboard}, by normalizing the infraction counts per meter. NS is defined as 

\begin{equation}
    \begin{aligned}
        \text{NS} = RC\cdot {P_I}^{IC}   \cdot {P_{RD}}
    \end{aligned}
\end{equation}
where $RC$ is route completion, $P_I$ represents the infraction penalty for collisions, $IC$ represents the number of robot collisions per meter, and $P_{RD}$ represents the penalty for route deviation.
Following the CARLA leaderboard settings, $P_I$ is set to $0.5$, and $P_{RD}$ is defined as
\begin{equation}
    P_{RD} = 
    \begin{dcases}
        0.8, &\text{if } {RD} > \epsilon_{RD} \\
        1.0, & \text{otherwise}
    \end{dcases}
\end{equation}
where $\epsilon_{RD}=1.5\text{m}$ is the route deviation threshold.

\boldparagraph{Energy Evaluation} 
In addition to the navigation-specific evaluations, we introduce metrics to assess the computational and communication overhead incurred during the execution of our algorithm. 

Specifically, the total energy consumption for each episode is defined as 
\begin{equation}
    \begin{aligned}
    \text{Energy}=E_{local} \cdot N_{local} + E_{cloud} \cdot N_{cloud} 
    \end{aligned}
\end{equation}
where $N_{local}$ and $N_{cloud}$ represent the number of steps the local and cloud models are chosen by the routing policy in one episode, respectively. $E_{local}$ and $E_{cloud}$ are the energy costs for processing one observation locally or on the cloud server.
We refer to the energy values suggested by~\cite{kang2017neurosurgeon} and scale them according to our model and image size. In our study, $E_{local}=0.15J$ is the computation energy cost of the local policy, and $E_{cloud}=1.5J$ comprises both the computation and communication energy costs of the cloud policy. Further details regarding the calculation can be found in the supplementary material.

\boldparagraph{Ecological Navigation Score} 
As our primary objective is to optimize the overall energy consumption while ensuring robot navigation performance, we introduce an \textit{Ecological Navigation Score}  (ENS) that balances navigation performance and efficiency considerations.

The ENS is defined as
\begin{equation}
    \begin{aligned}
        \text{ENS} =  {P_E} \cdot \text{NS}
    \end{aligned}
\end{equation}

where $P_E$ is the penalty term for energy consumption, and defined as 
\begin{equation}
    \begin{aligned}
         P_E = 1-\frac{\text{Energy}}{N_E}
    \end{aligned}
\end{equation}
where $N_E = \left(E_{local}+E_{cloud}\right)\cdot  \left(N_{local}+N_{cloud}\right)$ is a normalization factor. 

\boldparagraph{Run-Time} 
Real-time decision-making is crucial for robot navigation. In our study, the robots need to respond instantly to environmental changes and swiftly switch between local and cloud-based navigation policies to adjust path planning. To demonstrate this real-time response, we report our run-time Frames Per Second (FPS), which includes both the model processing time and the communication time between a local device and the cloud server.

\subsection{Results}
We use the CARLA simulator (version 0.9.13)~\cite{dosovitskiy2017carla} to validate the effectiveness of our proposed method.
We tested our algorithm on five different routes in CARLA's Town 10, each with 30 episodes. 
In testing time, we explored diverse weather conditions (including hard rain, sunny, wet, and sunset) and varying traffic density to simulate realistic conditions~\cite{codevilla2019exploring}. The maximum length of the routes is 40 meters.
In this section, we first compare UniLCD with state-of-the-art local-cloud collaboration solutions~\cite{yao2020deep,kang2017neurosurgeon,kag2022efficient,wang2019computation,van2021adaptive}.
Subsequently, we evaluate the performance of UniLCD with various sizes of local models and data-transmission settings. 
Third, we conduct ablation studies across different crowd-density settings. Finally, we systematically assess the significance of each reward term by omitting one at a time and analyzing their respective impacts, thereby reinforcing the foundational rationale of our reward design. 
The last two ablation studies can be found in our supplementary.

\colorlet{shadecolor}{gray!40}
\begin{table*}[t!]
    \normalsize
    \centering
	\caption{\textbf{Comparing UniLCD with Baselines.} We categorize our comparative analysis between individual models (Local and Cloud only), baselines (Deep Learning and RL-based computational offloading research), and our proposed UniLCD variations, which show progressive improvement. $\dag$ denotes methods transmitting to the cloud raw input data, \ie, instead of an embedding. ENS is Ecological Navigation Score (\%), NS is Navigation Score (\%), SR is Success Rate (\%), RC is Route Completion (\%), Infract. is Infraction Rate (/m), Energy is measured in Joules per meter (J/m) and FPS is Frames Per Second.} %  
    \resizebox{0.98\textwidth}{!}{%
    %@{~~~~~~~~~}
	\begin{tabular}{l|>{\columncolor[gray]{0.92}} c ccccccccc }	    
	    \textbf{Method}& 
	\multicolumn{1}{p{1.4cm}}{\centering\textbf{ENS}$\uparrow$} & \multicolumn{1}{p{1.4cm}}{\centering\textbf{NS}$\uparrow$} &
	\multicolumn{1}{p{1.4cm}}{\centering\textbf{SR}$\uparrow$} & \multicolumn{1}{p{1.4cm}}{\centering\textbf{RC}$\uparrow$}   &  \multicolumn{1}{p{1.4cm}}{\centering\textbf{Infract.}$\downarrow$}  & \multicolumn{1}{p{1.4cm}}{\centering\textbf{Energy}$\downarrow$}  & \multicolumn{1}{p{1.4cm}}{\centering\textbf{FPS}$\uparrow$}   &\\
    \toprule
    $\dag$ Cloud-Only~\cite{schneider2017regnet} &0.00 & 96.47 &93.33&98.50& 0.03 & 36.49 & 7.11& \\
    Local-Only~\cite{sandler2018mobilenetv2}&63.43 & 67.33 &0.00&75.23  & 0.16& 4.33 & 65.40 & \\
   
         \midrule
    \multicolumn{8}{l}{\underline{\textit{Baseline Methods:}}} \\
    % \midrule
     Compressive Offloading~\cite{yao2020deep} & 13.98 & 80.16 &0.00&80.16 &  0.00 & 90.66 & 1.82 & \\
     $\dag$ Selective Query~\cite{kag2022efficient}   &24.14 & 61.28 & {0.00} &82.68 & 0.11 & 45.35 & 18.14& \\
     $\dag$ Adaptive Offloading~\cite{van2021adaptive} &37.42 & 40.37 &70.00&94.05 &  1.22 & 4.80 & 30.14 & \\
     Neurosurgeon~\cite{kang2017neurosurgeon} &39.85 & 63.10 &0.00&{80.54}& {0.03} & {28.31} & 12.53 & \\
     SPINN~\cite{laskaridis2020spinn} &36.31 &72.75 &60.00 &92.73 &0.35 &18.94 & 20.37 & \\
        % Neurosurgeon~\cite{kang2017neurosurgeon} &31.55 & 63.10 &0.00&{80.54}& {0.03} & {28.31} & 12.53 & \\

 Deep Sequential RL~\cite{wang2019computation} &58.84 & 61.83 &0.00 & 79.36 & 0.36 & 3.77 & 77.94 \\
    \midrule
    \multicolumn{8}{l}{\underline{\textit{UniLCD Module Ablations:}}} \\
   % \midrule
      $\dag$ Standard Reward &48.35&54.99& 0.00&75.23 &  0.13 &  3.57 & 50.20& \\
      $\dag$  Standard Reward w/ History &50.04&57.21& 10.00&77.71 &  0.12 &  8.38 & 49.07& \\
      $\dag$  Our Reward (\cref{eq:reward}) &48.30&79.90& 56.66&{91.15} &  0.19&  21.72 & 16.05& \\
      $\dag$  Our Reward w/ History & 71.70 &87.71& 83.33&94.66 &  0.11 & 7.83& 33.98&\\
	  Our Reward (\cref{eq:reward}) &57.20&87.39& 60.00 &91.10 &  0.06 & 6.60&  \textbf{12.49}&\\
  % Our Reward w/ History &72.11&88.14& 90.00 &95.12&  0.11 & 5.79&  23.53&\\
   Our Reward w/ History &\textbf{85.97} &\textbf{94.58}& \textbf{93.33} &\textbf{95.90}&  \textbf{0.02} & \textbf{2.90}&  26.49&\\
   \bottomrule
	\end{tabular}%
}
\label{tab:s2rcompare}	
\end{table*}

% \boldparagraph{Comparing UniLCD with Baselines} 
\boldparagraph{Comparing UniLCD with Baselines} \cref{tab:s2rcompare} depicts our main results, comparing against multiple baselines, including local and cloud-only models. We emphasize that prior work does not usually consider safety-critical, real-time tasks such as social navigation. As shown in~\cref{tab:s2rcompare}, several baselines can achieve a viable route completion (RC) score, however, this comes at a cost, \eg, high infraction rates. This includes early-exit baselines, which provide a limited offloading mechanism. We show the ENS metric to drop to zero when assessing the high energy-consuming cloud model exclusively, demonstrating that conducting all computations in the cloud with the computation-heavy model is energy-consuming. Moreover, the local-only model demonstrates a poor navigation score of 67.33\% and a high collision rate of 0.16, indicating that low-accuracy, device-only models are more susceptible to navigation errors. The baseline models show comparable ENS scores to the local-only method but do not improve much in other metrics. Early-exit strategies such as SPINN~\cite{laskaridis2020spinn} and Neurosurgeon~\cite{kang2017neurosurgeon} address on-device efficiency with improved ENS scores up to 39.85\%, but perform worse than offloading-based methods in other metrics. Furthermore, we compare our method with DNN-based offloading (Selective Query~\cite{kag2022efficient}), which fails to improve due to the dynamic nature of the task and over-reliance on the cloud. This particular baseline also requires transmitting raw data to the cloud, which is inefficient. The best performing baseline, of Deep Sequential RL~\cite{wang2019computation}, achieves an ENS of 58.84\%.

UniLCD achieves a state-of-the-art results, with an ENS of 85.97\%, achieving state-of-the-art performance. Our analysis also highlights the need for cloud-edge collaborative methods that can combine both the strengths of fast local inference and high-capacity cloud computation.

\boldparagraph{Impact of Reward Design} \cref{tab:s2rcompare} also shows ablation of UniLCD using various rewards. We evaluate UniLCD using the standard reward design (\eg,~\cite{zhuang2023robot}), where individual reward terms are added. We initially find the additive reward to result in comparable performance to our reward design (ENS of 48.35\% compared to 48.30\%). However, when added history, the additive reward does not improve, while ours is shown to outperform significantly (ENS of 50.04\% for standard reward compared to 71.70\% with ours). We attribute this to the normalization effect of the multiplicative function during training. This eases optimization, whereas, in additive overall reward, the contribution of each term may need to be carefully optimized. 
While it is possible that careful hyperparameter optimization can match these results, this highlights the flexibility and generalization of our objective, which performs well across different model inputs. This finding also uncovers the sensitivity of current methods for the reward design. Finally, we utilize an embedding from the local model as input to UniLCD to provide the necessary context for collision handling and transmit it to the cloud, \ie, instead of raw data to reduce energy costs. This improves overall performance, achieving an ENS of 85.97\%, mostly due to reduced energy consumption, while maintaining high performance. The embedding-based communication architecture also reduces collisions due to a higher FPS, resulting in a higher NS.

\colorlet{shadecolor}{gray!40}
\begin{table*}[t!]
    \normalsize
    \centering
	\caption{\textbf{Local Policy Backbone Ablations.} We test the task performance in the dense crowd setting for UniLCD and UniLCD w/ History. $\dag$ denotes methods transmitting to the cloud raw input data, \ie, instead of an embedding. For UniLCD transmitting data, we use three dimensions of local models: Tiny, Small, and Medium. For UniLCD transmitting embedding, we use local models trained up to two stages: Stage 1 trained with fewer parameters and Stage 2 trained with more parameters. Params is Number of Parameters in each local model (M), ENS is Ecological Navigation Score (\%), NS is Navigation Score (\%), SR is Success Rate (\%), RC is Route Completion (\%), Infract. is Infraction Rate (/m), Energy is measured in Joules per meter (J/m), and FPS is Frames Per Second.} 
    \resizebox{0.98\textwidth}{!}{%
 \begin{tabular}{l | c |  >{\columncolor[gray]{0.92}} c ccccccccc}
	    \textbf{Local Model Size}& 
	\multicolumn{1}{p{1.4cm}}
 {\centering\textbf{Params}} & \multicolumn{1}{p{1.4cm}}{\centering\textbf{ENS}$\uparrow$} & \multicolumn{1}{p{1.4cm}}{\centering\textbf{NS}$\uparrow$} &
	\multicolumn{1}{p{1.4cm}}{\centering\textbf{SR}$\uparrow$} & \multicolumn{1}{p{1.4cm}}{\centering\textbf{RC}$\uparrow$}   &  \multicolumn{1}{p{1.4cm}}{\centering\textbf{Infract.}$\downarrow$}  & \multicolumn{1}{p{1.4cm}}{\centering\textbf{Energy}$\downarrow$}  & \multicolumn{1}{p{1.4cm}}{\centering\textbf{FPS}$\uparrow$}   &\\
    \toprule
    \multicolumn{8}{l}{\underline{\textit{UniLCD:}}} \\
    % \midrule
        $\dag$ Tiny  &1.37&12.80   &93.29&{90.00} & 97.93 &  0.07 & 34.01 & 7.35 \\
        $\dag$ Small &2.54&48.30   &79.90&{56.66} &91.15& 0.19 & 21.72&  16.05&\\
        $\dag$ Medium &3.50&50.92  &87.87&{80.00} & 95.50 & 0.12 & 21.19 &15.10& \\
    \midrule
    \multicolumn{8}{l}{\underline{\textit{UniLCD w/ History:}}} \\
   % \midrule
$\dag$ Tiny &1.37&0.00    &91.27&\textbf{93.33}  & \textbf{98.50}  & 0.11& 36.52 &6.42&  \\
       $\dag$ Small &2.54&71.70   &87.71& {83.33} &94.66  & 0.11 & 7.83&  33.98&\\
$\dag$ Medium  &3.50&73.46  &83.86&  90.00 & 96.22& 0.15&5.22 & 11.53& \\
\bottomrule
    \multicolumn{8}{l}{\underline{\textit{UniLCD:}}} \\
    Stage 1&0.53&57.20    &87.39&60.00 & 91.10 & 0.06 & 6.60 & 12.49& \\
    Stage 2&0.95&74.12  &81.54& 93.33 & 91.10 & 0.16 & 1.80 &\textbf{65.40}& \\
    \midrule
    \multicolumn{8}{l}{\underline{\textit{UniLCD w/ History:}}} \\
   % \midrule
      Stage 1 &0.53&85.97     &94.58&\textbf{93.33} & 95.99  & \textbf{0.02}& 2.90 & 26.49& \\
        Stage 2 &0.95&\textbf{86.78}   &\textbf{95.47}& \textbf{93.33} & \textbf{98.15} & 0.04& \textbf{1.77} &36.50& \\
   \bottomrule
	\end{tabular}%
}
\label{tab:s2rlocal}	
\end{table*}

\boldparagraph{Local Policy Backbone Ablations}
\cref{tab:s2rlocal} shows ablations of UniLCD for local backbones of various sizes. We run experiments with UniLCD that transmit raw image data to the cloud using local backbones such as MobileNetV2 (medium), MobileNetV3small (small), and MobileVIT (tiny), coupled with a RegNet (cloud) model. As an imitation model, the medium backbone demonstrates task performance closest to the cloud, with minimal instances of infractions followed by the small and tiny backbones. The UniLCD with history model consistently favors local at instances where both local and cloud models have similar capacity, \ie, the medium backbone. The algorithm successfully sustains an impressive average route completion rate of 94.66\% with occasional cloud support and significantly reduces energy consumption to 7.83 J/m using a comparatively weaker local model with a smaller backbone. Further, testing local policy incapable of path-following or collision avoidance with a tiny backbone reveals a complete reliance on cloud assistance but results in the successful completion of all episodes. 

We also run ablations of UniLCD that transmit embeddings to the cloud. The local models for these experiments are trained by freezing pre-trained cloud weights where stage 1 is fine-tuned with the feature extractor containing the early layers of the cloud model and stage 2 is fine-tuned with the feature extractor containing the final layers of the cloud model. Observations indicate that UniLCD with history model achieves an ENS of 85.97\% solely with the stage 1 backbone. It is noteworthy that both models exhibit a bias towards local processing when a backbone comparable to cloud performance (stage 2) is employed, resulting in the highest ENS of 86.78\%, and a route completion of 98.15\% with minimal infractions, averaging 0.04 per meter. Overall, these findings underscore UniLCD's proficiency in leveraging cloud resources only when local processing is inadequate.
\begin{figure}[t]
    \centering
    % \begin{tabular}{c}
    \includegraphics[trim={0 0 0 0.2 cm},clip,width=0.5\textwidth]{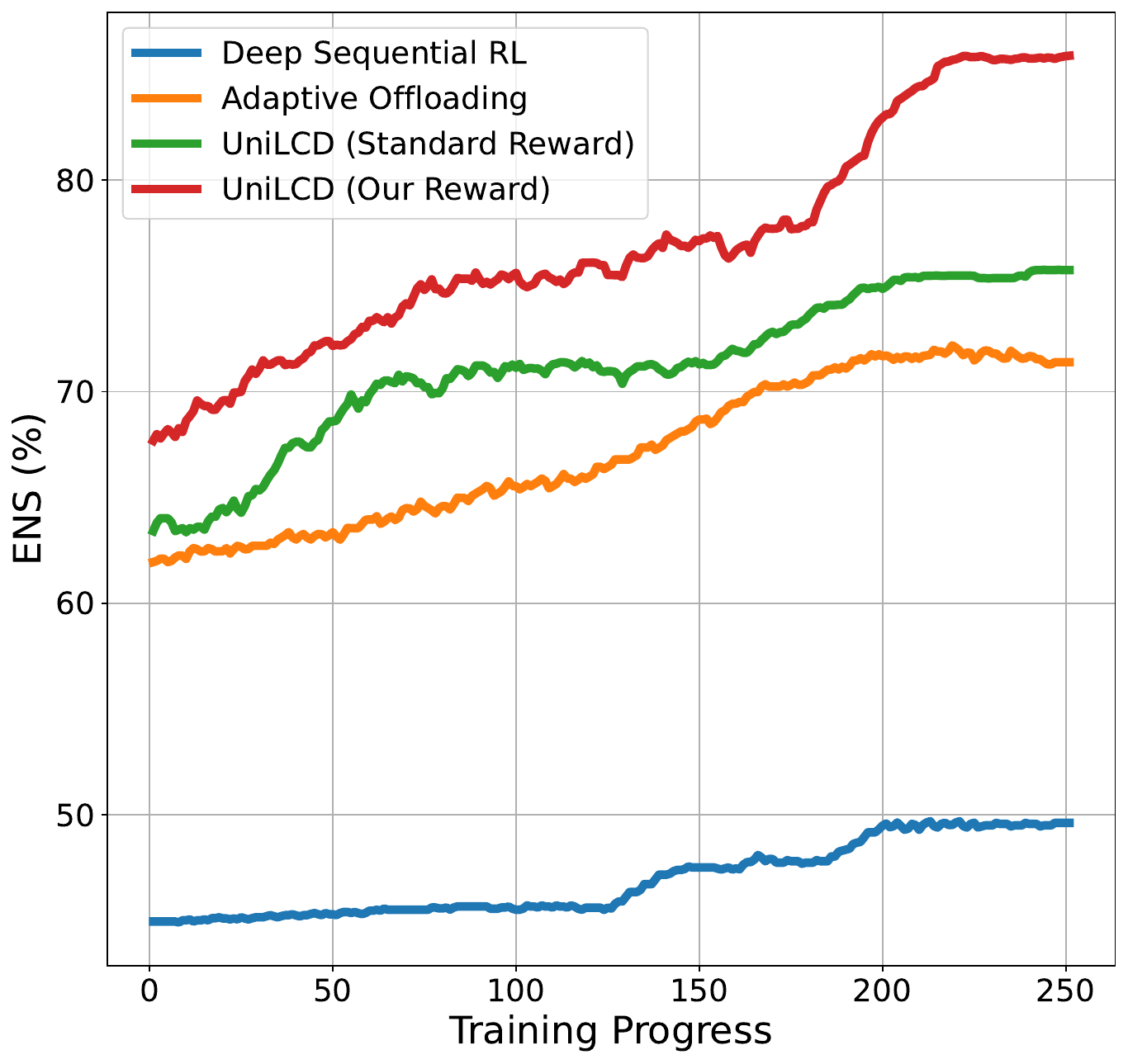}
    \caption{\textbf{Training Progress Results.} We evaluate performance for different models, including UniLCD trained with a standard, carefully tuned, additive reward vs. UniLCD with the proposed reward function. We find consistently improved performance throughout the entire model training process. Results are shown for averaging across 10 evaluation seeds. 
    }
    \label{fig:ENSPlot}
\end{figure}

\boldparagraph{Performance over Training Progress} We also investigate the evolution of the ENS task metric over the training progress as shown in~\cref{fig:ENSPlot}. Remarkably, our model and reward outperform the standard reward even at the onset of training despite not requiring careful hyper-parameter tuning. Nonetheless, as training progresses, the difference in the baselines becomes even more apparent. The ENS performance is presented over training iterations and averaged over 10 tests, \ie, using different environmental seeds. Results for performance in different environments and settings (\eg, crowd density) can be found in the supplementary. In our comparative analysis, we note that as the complexity of the scenario increases, our baseline approaches show a decline in task performance. This decline is evidenced by an increase in route deviation and infraction rates. While UniLCD can easily navigate comparatively simpler scenarios, its infraction rate increases only slightly. This observation suggests the algorithm's robust adaptability for effective navigation in diverse and intricate environments.
\section{Conclusion and Future Work}
% \vspace{0.2cm}
We envision large-scale robot agents that collaborate and actively execute tasks in complex real-world scenarios. To this end, we design a PPO-based routing policy that learns to seamlessly switch between local and cloud policies depending on the situation and task objective of the observations. Specifically, we address the challenging task of real-time social robot navigation. We demonstrate that the proposed approach results in reduced overall energy cost while maintaining robust navigation performance. Our simulation environment can be used to facilitate more research into both \textit{effective and efficient} navigation in the future. Thus, we aim to facilitate the development of more sustainable and robust societal-scale AI-based systems. Moreover, our approach can be extended to various platforms and real-time decision-making tasks that require both seamless communication and high performance. While we take a step towards quantifying trade-offs in real-time, safety-critical systems in challenging simulation settings, the next step would be to analyze UniLCD within diverse real-world environments and ambient settings (\eg, different communication settings, larger cloud models). Finally, although we consider the overarching objective of minimizing processing computation across both local and cloud devices, other aspects of efficiency can also be considered. For instance, maintaining local infrastructure often requires additional battery and hardware resources, which can be better optimized with cloud-based solutions especially when leveraging large, high-capacity models~\cite{cheng2018model,dong2024creating}. 
\label{sec:conclusion}
\section*{Acknowledgments}
We thank the Red Hat Collaboratory (award \#2024-01-RH07) for supporting this research, and Bassel Mabsout for helpful discussions regarding the problem formulation.

\bibliographystyle{splncs04}
\bibliography{egbib}

\clearpage
\onecolumn
\begin{center}
%Please do not change title, create a seperate .tex file for supp
      {\Large \bf %Supplementary Material for \\
      Supplementary for Unified Local-Cloud Decision-Making via \\ Residual Reinforcement Learning \par}
\end{center}

\setcounter{section}{0}
\begin{abstract}
This supplementary provides additional implementation and ablative details. Specifically, we discuss additional details regarding the introduced cloud and energy-aware environment 
% (Sec.~\ref{sec:imp}) 
as well as additional ablative analysis. 
% (Sec.~\ref{sec:res}). 
Our supplementary video depicts qualitative results, comparing roll-outs generated by the proposed model to baseline models. 
\end{abstract}

\section{Environment Implementation}
\label{sec:imp}

\boldparagraph{Energy Cost Model} To quantify energy consumption, we assume a standard cost per floating point operation model~\cite{kang2017neurosurgeon,hadidi2019characterizing,li2024intelligent}. For instance, a 675 Kb $480\times480\times3$ image and a 1.37 M parameters MobileVIT~\cite{mehta2021mobilevit} model would result in energy consumption on a local GPU that is 0.15 J (based on a factor of 0.095 J per flop~\cite{kang2017neurosurgeon}). 
We further add an energy cost for every communication, \ie, transmitting raw data or an embedding to the cloud~\cite{kag2022efficient}. Here, energy consumed for local-cloud communication is computed as $6.94\times{10^{-5}}$ J per byte~\cite{kang2017neurosurgeon}. When transmitting only an embedding (a $24\times24$ array) from the local policy as an intermediate input to the cloud policy, the communication energy consumption decreases, \eg, from the original image which requires 1.55 J to 25.18 mJ.

\boldparagraph{Cloud Communication Model} Cloud transmission involves a latency, which the model should optimize over, \eg, in safety-critical scenarios based on the task. Our communication model resembles the one used by Kang~etal~\cite{kang2017neurosurgeon}, where latency is modeled as a Gaussian with an average time of 0.5 s and variance of 0.1 s. To extend our research, we also experimented with modeling latency using a Pareto distribution. Under the Pareto distribution, we observed a 7\% reduction in ENS, with an ENS value of 80.45, infractions at 0.03 /m, and energy consumption at 2.76 J/m. While this is a conservative estimate, we will release our simulation, code, and models to facilitate the analysis of RL agents across diverse computational and latency configurations.  

\subsection{Reward Design}

\begin{figure}[t]
    \centering
    \includegraphics[trim={0 0 0 0.2 cm},clip,width=0.7\textwidth]{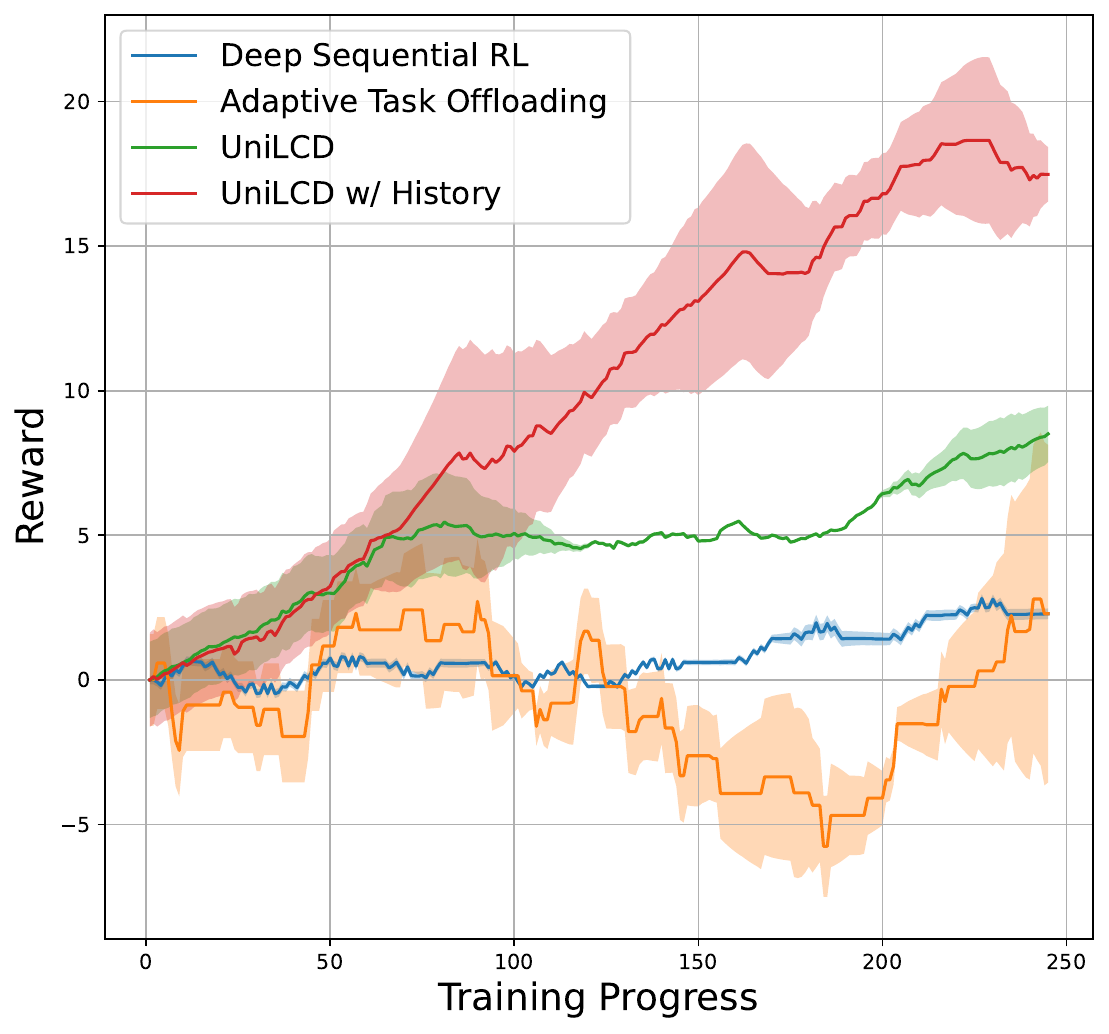}
    \caption{\textbf{Training Progression.} We evaluate the rewards for our overall framework throughout the training process against baseline models. We demonstrate high sample-efficiency compared to prior methods, particularly when the routing module is given a history input, \ie, prior actions and their source (cloud or local decision). Despite common instabilities in training reinforcement learning models, UniLCD is shown to achieve significantly higher reward early in the training process. }
    \label{fig:rewardPlot}
    % \vspace{0.2cm}
\end{figure}

\boldparagraph{Reward Progression} 
 Fig. \ref{fig:rewardPlot} shows our reward analysis for RL baselines and UniLCD models. The reward progression indicates that the baselines fail to learn any meaningful information throughout the training and continue to oscillate within a very narrow range. Our UniLCD demonstrates a slight improvement in the initial episodes but exhibits delayed convergence due to its slower learning rate. Remarkably, the UniLCD with history model displays a distinct pattern of rapid advancement, characterized by a steep rise in the initial episodes. Consequently, it is clear that this configuration learns and adapts to the desired behavior more efficiently than the comparative approaches, even within a few iterations. The primary benefits in sample efficiency stem from the definition of the reward function definition and the careful design of the routing module and its inputs.
 
\boldparagraph{Comparison with Standard Additive Reward} In the main paper, we find it crucial to carefully shape the reward function in order to effectively optimize for several different types of rewards. Our reward is contrasted with current standard approaches that linearly combine the rewards~\cite{zhuang2023robot,booth2023perils,knox2023reward,ke2024real}, \ie, 
\begin{equation}
\begin{aligned}
%r_{standard}=\alpha\left({r_{geo}+r_{speed}+r_{energy}}\right) -r_{collision}
r_{standard}=\alpha_g{r_{geo}+\alpha_s r_{speed}+\alpha_e r_{energy}}+\alpha_c r_{collision}
\end{aligned}
\label{eqn:add}
\end{equation}
adding the geodesic, speed, energy, and collision rewards and $\alpha_g$, $\alpha_s$, $\alpha_e$, $\alpha_c$, are tuned scalar hyperparameters (we perform careful tuning of these with a grid search). 

\colorlet{shadecolor}{gray!40}
\begin{table*}[t!]
    % \normalsize
    \centering
	\caption{\textbf{Reward Component Ablation.} We investigate the impact of each reward component in our reward design for UniLCD.  
 ENS is Ecological Navigation Score (\%), NS is Navigation Score (\%), SR is Success Rate (\%), RC is Route Completion (\%), Infract. is Infraction Rate (/m), Energy is measured in Joules per meter (J/m), and FPS is Frames Per Second (empirically measured and averaged throughout the trials).} 
 \label{tab:reward}	
 %$\dag$ denotes router module with history as input.   }
    \label{tab:rewardcomp}
	% \vspace{-0.2cm}
    \resizebox{0.98\textwidth}{!}{%
    %@{~~~~~~~~~}
	% \begin{tabular}{l|>{\columncolor[gray]{0.92}} c cccccccccc }	
 \begin{tabular}{l | >{\columncolor[gray]{0.92}} c ccccccccc}
	    \textbf{Reward}&   \multicolumn{1}{p{1.4cm}}{\centering\textbf{ENS}$\uparrow$} & \multicolumn{1}{p{1.4cm}}{\centering\textbf{NS}$\uparrow$} &
	\multicolumn{1}{p{1.4cm}}{\centering\textbf{SR}$\uparrow$} & \multicolumn{1}{p{1.4cm}}{\centering\textbf{RC}$\uparrow$}   &  \multicolumn{1}{p{1.4cm}}{\centering\textbf{Infract.}$\downarrow$}  & \multicolumn{1}{p{1.4cm}}{\centering\textbf{Energy}$\downarrow$}  & \multicolumn{1}{p{1.4cm}}{\centering\textbf{FPS}$\uparrow$}   &\\
    \toprule
       All Terms &\textbf{85.97}   &\textbf{94.58}& \textbf{93.33} & \textbf{95.90} &  \textbf{0.02} & \textbf{2.90}& \textbf{26.49} \\
        w/o $r_{geo}$ &67.04 &74.65&0.00 & 76.22 & 0.03 & 7.42 &58.82& \\
         w/o $r_{speed}$ &66.05  &72.66&0.00 &77.34& 0.09 & 6.61&  65.40&\\
          w/o $r_{energy}$ &0.00  &93.53&90.00 & 95.50 & 0.03 & 43.75 &6.42& \\
    \bottomrule
	\end{tabular}%
}
\end{table*}

\boldparagraph{Impact of Reward Components} Table~\ref{tab:reward} analyzes the impact of various reward terms on task performance. We show each term to contribute to the overall task performance. We note that we do not multiply the collision reward with the other terms in order to emphasize it during training. This provides a \textbf{major negative disadvantage to avoid collisions}, as it dominates the overall reward once it occurs (and results in episode termination). Moreover, we show that without penalizing energy consumption of communication to and processing in the cloud, the model effectively learns to rely on the more powerful model. Finally, while the imitation-learned models take as input the goal, adding a geodesic reward term is found to help ensure the routing module and final action prediction do not result in a policy that deviates from the planned path. 

\boldparagraph{Real-world Validation} Our simulation experiments are based on real-world network latency calculations, as detailed in Sec~\ref{sec:imp}. To extend our research into real-world validations, we conducted a small-scale experiment using an RC vehicle platform (Traxxas X-Maxx) equipped with a Jetson Nano. In this setup, reward terms were computed in a self-supervised manner using an EAI XL2 LiDAR with a minimum range of 0.1 meters. Given the need for high sample efficiency in real-world reinforcement learning (RL), we performed an ablation study comparing Proximal Policy Optimization (PPO) and CrossQ. In Line 16 of Algorithm 1, CrossQ learns a policy in approximately five minutes, demonstrating a 21\% improvement over [31].

\boldparagraph{Further Work} While our work takes a step towards developing vision-based systems and effectively address multifaceted objectives, several simplifications were necessary. Due to the difficulty and sample inefficiency associated with the optimization of RL agents for energy, latency, and safety constraints -- all are crucial in real-world settings -- we adopted a step-by-step training approach. Specifically, the local and cloud policies were trained using imitation learning by assuming expert trajectories. Our UniLCD pipeline can readily support additional model fine-tuning and incorporation of consumption-based reward terms, \eg, both the cloud and local models can be fine-tuned using RL, yet this requires additional study in the future. 

\colorlet{shadecolor}{gray!40}
\begin{table*}[t!]
    % \normalsize
    \centering
	\caption{\textbf{Pedestrian Density Ablation.} Model performance is shown for different pedestrian densities along the path, from Low (mostly empty, 5 pedestrians), Medium (15 pedestrians), High (30 pedestrians), and up to Crowded (70 pedestrians). $\dag$ denotes methods transmitting to the cloud raw input data instead of an embedding. ENS is Ecological Navigation Score (\%), NS is Navigation Score (\%), SR is Success Rate (\%), RC is Route Completion (\%), Infract. is Infraction Rate (/m), Energy is measured in Joules per meter (J/m), and FPS is Frames Per Second.}
 
    \label{tab:baseline}
	% \vspace{-0.2cm}
    \resizebox{0.90\textwidth}{!}{%
    %@{~~~~~~~~~}
	\begin{tabular}{l|l|>{\columncolor[gray]{0.92}} c ccccccccc }	    
	    \textbf{Method}& \textbf{Setting}& 
	\multicolumn{1}{p{1.4cm}}{\centering\textbf{ENS}$\uparrow$} & \multicolumn{1}{p{1.4cm}}{\centering\textbf{NS}$\uparrow$} &
	\multicolumn{1}{p{1.4cm}}{\centering\textbf{SR}$\uparrow$} & \multicolumn{1}{p{1.4cm}}{\centering\textbf{RC}$\uparrow$}   &  \multicolumn{1}{p{1.4cm}}{\centering\textbf{Infract.}$\downarrow$}  & \multicolumn{1}{p{1.4cm}}{\centering\textbf{Energy}$\downarrow$}  & \multicolumn{1}{p{1.4cm}}{\centering\textbf{FPS}$\uparrow$}   &\\
    \toprule
    \multirow{4}{*}{$\dag$ Cloud-Only~\cite{radosavovic2020designing}} 
    & Low &0.00 & 98.41 &95.00&99.10& 0.01 & 33.93 & 8.91& \\
    & Medium &0.00 & 97.14 &93.33&98.50& 0.02 & 35.11 & 8.74& \\
    & High &0.00 & 96.47 &93.33&98.50& 0.03 & 36.49 & 7.11& \\
    & Crowd &0.00 & 88.24 &92.00&95.24& 0.11 & 48.93 & 6.52& \\
\midrule
    \multirow{4}{*}{Local-Only~\cite{sandler2018mobilenetv2}}
    & Low &66.18 &69.70 &0.00&75.23  & 0.11& 3.78 & 78.19 & \\
    & Medium &65.99 & 69.70 &0.00&75.23  & 0.11& 3.98 & 73.67 & \\
    & High &63.43 & 67.33 &0.00&75.23  & 0.16& 4.33 & 68.40 & \\
    & Crowd &35.47 & 37.55 &0.00&63.16  & 0.75& 4.94 & 65.40 & \\
        \bottomrule
    \multicolumn{9}{l}{\textit{Baseline Methods:}} \\
    \toprule
    % \midrule
    \multirow{4}{*}{SPINN~\cite{laskaridis2020spinn}}
    & Low &43.71 & 85.13 &70.00&{95.12}& {0.16} & {16.88} & 23.84 & \\
    & Medium &39.94 & 80.60 &66.66&{93.24}& {0.21} & {16.74} & 24.21 & \\
    & High &36.31 & 72.75 &60.00&{92.73}& {0.35} & {18.94} & 20.37 & \\
    & Crowd &32.07 & 53.22 &60.00&{90.76}& {0.77} & {17.16} & 21.84 & \\
    
    \hline

    \multirow{4}{*}{Compressive Offloading~\cite{yao2020deep}} 
    & Low & 11.81 & 80.43 &0.00&80.43 &  0.00 & 88.23 & 2.46 & \\
    & Medium & 12.38 & 80.16 &0.00&80.16 &  0.00 & 89.12 & 2.17 & \\
    & High & 13.98 & 80.16 &0.00&80.16 &  0.00 & 90.66 & 1.82 & \\
    & Crowd & 10.06 & 75.66 &0.00&75.66 &  0.00 & 92.66 & 1.19 & \\

    \hline
    
    \multirow{4}{*}{$\dag$ Selective Query~\cite{kag2022efficient}}
    & Low &27.87 & 64.78 & {0.00} &82.68 & 0.03 & 42.64 & 21.57& \\
    & Medium &25.40 & 60.63 & {0.00} &80.12 & 0.08 & 44.87 & 19.39& \\
    & High &24.14 & 61.28 & {0.00} &82.68 & 0.11 & 45.35 & 18.14& \\
    & Crowd &14.90 & 44.27 & {0.00} &79.36 & 0.52 & 51.72 & 14.38& \\

    \hline 

    \multirow{4}{*}{Neurosurgeon~\cite{kang2017neurosurgeon}}
    & Low &43.79 & 67.62 &0.00&{85.12}& {0.01} & {25.62} & 12.53 & \\
    & Medium &42.12 & 65.12 &0.00&{82.54}& {0.02} & {26.48} & 13.21 & \\
    & High &39.85 & 63.10 &0.00&{80.54}& {0.03} & {28.31} & 12.53 & \\
    & Crowd &32.94 & 51.19 &0.00&{75.58}& {0.24} & {29.18} & 9.61 & \\

    \hline

    \multirow{4}{*}{$\dag$ Adaptive Offloading~\cite{van2021adaptive}}
    & Low &89.38 & 97.15 &90.0&98.51 &  0.02 & 5.02 & 34.91 & \\
    & Medium &51.87 & 55.56 &80.00&97.42 &  0.81 & 4.22 & 33.39 & \\
    & High &37.42 & 40.37 &70.00&94.05 &  1.22 & 4.80 & 30.14 & \\
    & Crowd &33.96 & 37.67 &80.00 &95.38 &  1.34 & 6.39 & 28.71 & \\

    \hline

    \multirow{4}{*}{Deep Sequential RL~\cite{wang2019computation}}
    & Low &70.15 & 73.02 &0.00 & 79.36 & 0.12 & 3.07 & 78.10 \\
    & Medium &66.66 & 69.56 &0.00 & 79.36 & 0.19 & 3.25 & 78.03 \\
    & High &58.84 & 61.83 &0.00 & 79.36 & 0.36 & 3.77 & 77.94 \\
    & Crowd &26.47 & 27.93 &0.00 & 75.28 & 1.43 & 4.30 & 76.12 \\
       
         \bottomrule
    \multicolumn{9}{l}{\textit{UniLCD Variations:}} \\
    \toprule
   
    \multirow{4}{*}{$\dag$ Standard Reward}
    & Low &49.37&56.15& 0.00&75.23 &  0.10 &  2.47 & 52.63& \\
    & Medium &49.37&56.15& 0.00&75.23 &  0.10 &  3.12 & 51.18& \\
    & High &48.35&54.99& 0.00&75.23 &  0.13 &  3.57 & 50.20& \\
    & Crowd &30.60&34.80& 0.00&75.23 &  0.79 &  4.86 & 47.74& \\

    \hline
    
    \multirow{4}{*}{$\dag$  Standard Reward w/ History}
    & Low &51.44&58.81& 10.00&77.71 &  0.08 &  7.02 & 51.42& \\
    & Medium &50.38&57.60& 10.00&77.71 &  0.11 &  7.57 & 50.20& \\
    & High &50.04&57.20& 10.00&77.71 &  0.12 &  8.38 & 49.07& \\
    & Crowd &35.22&40.26& 10.00&75.23 &  0.58 &  10.57 & 43.16& \\

    \hline
    
    \multirow{4}{*}{$\dag$  Our Reward}
    & Low &51.82&85.73& 60.00&{90.00} &  0.07&  18.49 & 20.82& \\
    & Medium &51.20&84.71& 60.00&92.70 &  0.13 & 20.57 & 17.14& \\
    & High &48.30&79.90& 56.66&{91.15} &  0.19&  21.72 & 16.05& \\
    & Crowd &38.79&64.16& 60.00&86.45 &  0.43 &  24.57 & 11.20& \\

    \hline
    
    \multirow{4}{*}{$\dag$  Our Reward w/ History}
    & Low & 73.72 &90.17& 83.33&94.66 &  0.07 & 5.78& 36.61&\\
    & Medium & 72.20 &88.32& 83.33&94.66 &  0.10 & 6.51& 34.73&\\
    & High & 71.70 &87.71& 83.33&94.66 &  0.11 & 7.83& 33.98&\\
    & Crowd & 59.04 &72.22& 83.33&93.33 &  0.37 & 10.83& 30.62&\\

    \hline
	  
    \multirow{4}{*}{Our Reward}
    & Low &61.88&94.54& 60.00 &95.20 &  0.01 & 5.51&  \textbf{14.18}&\\
    & Medium &60.68&92.71& 60.00 &94.66 &  0.03 & 5.88&  \textbf{13.72}&\\
    & High &57.20&87.39& 60.00 &91.10 &  0.06 & 6.60&  \textbf{12.49}&\\
    & Crowd &50.22&76.74& 60.00 &90.00 &  0.23 & 8.43&  \textbf{10.73}&\\
  % Our Reward w/ History &72.11&88.14& 90.00 &95.12&  0.11 & 5.79&  23.53&\\

    \hline
   
    \multirow{4}{*}{Our Reward w/ History}
    & Low &\textbf{88.10} &\textbf{96.92}& \textbf{93.33} &\textbf{97.60}&  \textbf{0.01} & \textbf{2.58}&  30.02&\\
    & Medium &\textbf{87.06} &\textbf{95.77}& \textbf{90.00} &\textbf{96.44}&  \textbf{0.01} & \textbf{2.63}&  29.12&\\
    & High &\textbf{85.97} &\textbf{94.58}& \textbf{93.33} &\textbf{95.90}&  \textbf{0.02} & \textbf{2.90}&  26.49&\\
    & Crowd &\textbf{83.04} &\textbf{91.36}& \textbf{93.33} &\textbf{95.90}&  \textbf{0.07} & \textbf{3.12}&  22.84&\\
   
   \bottomrule
	\end{tabular}%
}
\label{tab:stochasticcomparefull}	
\end{table*}

\subsection{Performance Over Task Difficulty}
\begin{figure*}[!t]
    \centering
    % \vspace{0cm}
    \includegraphics[trim={0cm 0cm 0cm 0cm},width=\textwidth]{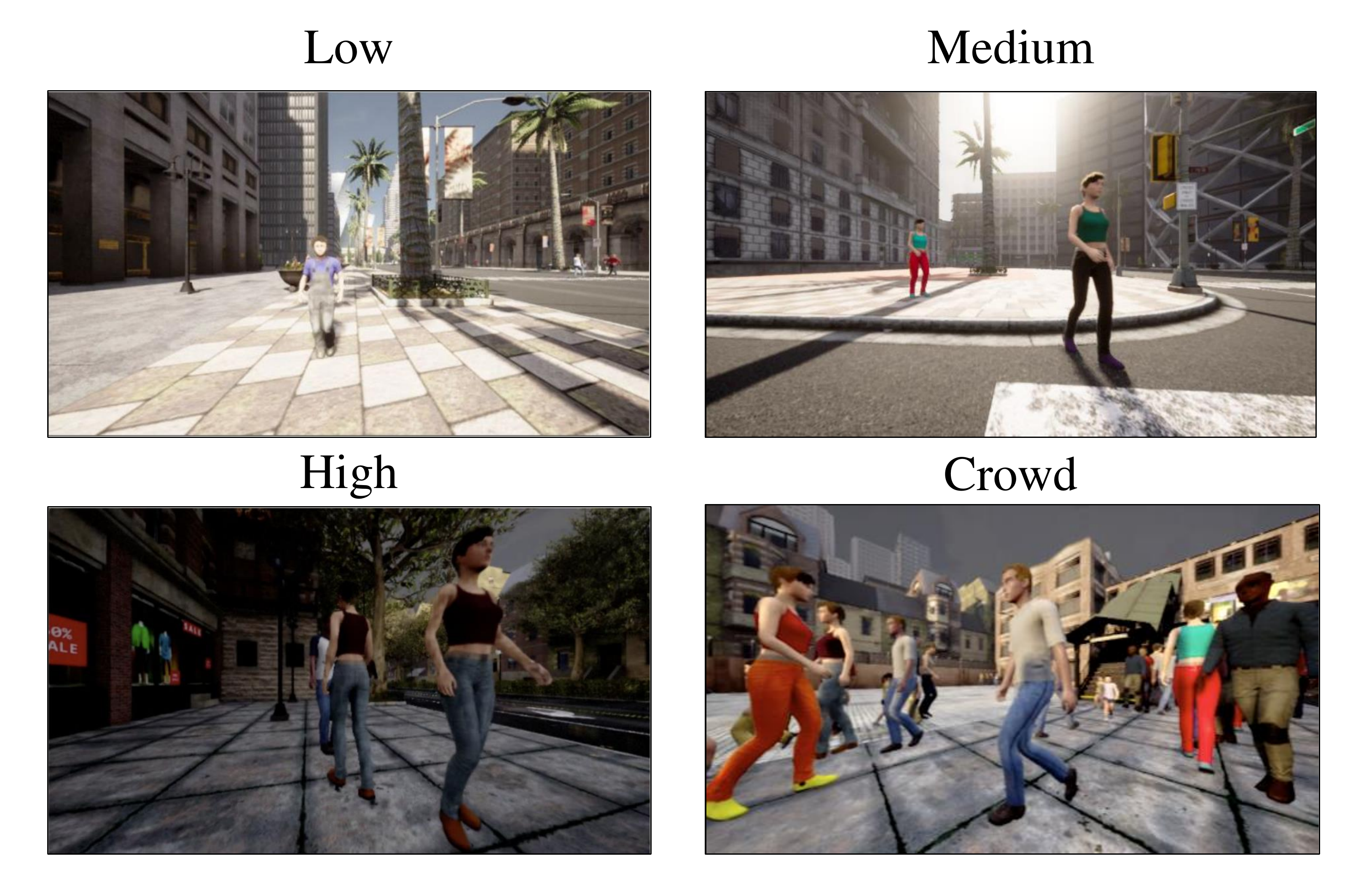} % Adjust width as necessary for two columns
    \caption{\textbf{Examples of Different Environmental Settings in Our Robot Navigation Environment.} We vary the pedestrian density in order to stress-test the proposed UniLCD method. Pedestrian count along the path ranges from 5 (Low), 15 (Medium), 30 (High), and 70 (Crowd). 
    }
    \label{fig:data}
\end{figure*}

The results in the main paper are shown for the high (30 pedestrians) density settings of our environment, spawned along the route (Fig.~\ref{fig:data}). For completeness, we report in Table~\ref{tab:stochasticcomparefull} the model performance across additional pedestrian density settings along the path, starting from a low-density scenario with 5 pedestrians to a crowded scenario with 70 pedestrians. As shown in the table, several models degrade in performance, particularly in crowded settings. Remarkably, our final UniLCD model exhibits only slight degradation (1\%), in terms of ENS, in the most challenging and crowded settings.

\clearpage

\end{document}